\documentclass[sigconf]{acmart}
\renewcommand\footnotetextcopyrightpermission[1]{} 
\setcopyright{none}

\def\BibTeX{{\rm B\kern-.05em{\sc i\kern-.025em b}\kern-.08emT\kern-.1667em\lower.7ex\hbox{E}\kern-.125emX}}

\usepackage{multirow}
\usepackage{array}
\usepackage{siunitx}
\usepackage{nicefrac}
\usepackage{graphicx}
\usepackage{mwe}
\usepackage{subcaption}
\newcolumntype{L}[1]{>{\raggedright\let\newline\\\arraybackslash\hspace{0pt}}m{#1}}
\newcolumntype{C}[1]{>{\centering\let\newline\\\arraybackslash\hspace{0pt}}m{#1}}
\newcolumntype{R}[1]{>{\raggedleft\let\newline\\\arraybackslash\hspace{0pt}}m{#1}}

\DeclareMathOperator*{\argmax}{argmax}

\begin{document}

\title{"Does 4-4-2 exist?" -- An Analytics Approach to Understand and Classify Football Team Formations in Single Match Situations}

\author{Eric M\"uller-Budack}
\email{eric.mueller@tib.eu}
\orcid{0000-0002-6802-1241}
\affiliation{%
 \institution{Leibniz Information Centre for Science and Technology~(TIB)}
 \city{Hannover}
 \state{Germany}
}

\author{Jonas Theiner}
\email{theiner@stud.uni-hannover.de}
\affiliation{%
 \institution{Leibniz Universit\"at Hannover}
 \city{Hannover}
 \state{Germany}
}

\author{Robert Rein}
\email{r.rein@dshs-koeln.de}
\affiliation{%
 \institution{German Sport University Cologne}
 \city{Cologne}
 \state{Germany}
}

\author{Ralph Ewerth}
\email{ralph.ewerth@tib.eu}
\orcid{0000-0003-0918-6297}
\affiliation{%
 \institution{L3S Research Center, Leibniz Universit\"at Hannover \& Leibniz Information Centre for Science and Technology~(TIB)}
 \city{Hannover}
 \state{Germany}
}

\renewcommand{\shortauthors}{Eric M\"uller-Budack, Jonas Theiner, Robert Rhein, Ralph Ewerth}

\begin{abstract}
The chances to win a football match can be significantly increased if the right tactic is chosen and the behavior of the opposite team is well anticipated. For this reason, every professional football club employs a team of game analysts. However, at present game performance analysis is done manually and therefore highly time-consuming. Consequently,  automated tools to support the analysis process are required. In this context, one of the main tasks is to summarize team formations by patterns such as 4-4-2. 
In this paper, we introduce an analytics approach that automatically classifies and visualizes the team formation based on the players' position data. We focus on single match situations instead of complete halftimes or matches to provide a more detailed analysis. 
The novel classification approach calculates the similarity based on pre-defined templates for different tactical formations.  
A detailed analysis of individual match situations depending on ball possession and match segment length is provided. For this purpose, a visual summary is utilized that summarizes the team formation in a match segment. An expert annotation study is conducted that demonstrates 1)~the complexity of the task and 2)~the usefulness of the visualization of single situations to understand team formations. The suggested classification approach outperforms existing methods for formation classification. In particular, our approach gives insights about the shortcomings of using patterns like 4-4-2 to describe team formations.
%
\end{abstract}

%

\maketitle
%
%
%
\section{Introduction}
\label{chp:intro}
The choice of the right tactic in a football match can have a decisive influence on the result and thus has a great impact on the success of professional football clubs~\cite{Rein2016}. According to \citet{garganta2009trends} and \citet{fradua2013designing}, the tactic describes how a team manages space, time and individual actions during a game. To select an appropriate tactic, detailed analyses are necessary to reveal and eventually exploit insights of the opposite team's behavior and patterns. These decisions are usually left to domain experts such as the coaching, analysts and scouting staff who observe and analyze entire football matches in order to prepare the next match. However, this process is very time-consuming which has limited its application in the past~\cite{james2006role,Rein2016}. For this reason, as the amount of available game performance data is steadily increasing the demand for automated analysis tools to support the scouting process is rapidly growing. 

Early approaches are mainly based on the interpretation of match statistics such as the distribution of the ball possession as well as shot, pass and tackle variables with the general aim to predict successful teams~\cite{oberstone2009differentiating, lago2010performance, jankovic2011influence, jones2004possession, redwood2012impact, pelechrinis2016sportsnetrank, cintia2015network}.
But, these statistics discard most contextual information as they are usually calculated across extended game periods like halftimes, whole matches or seasons. Therefore, these measures are not able to capture the increasing complexity of tactic in modern football and lack explanatory power in terms of prediction variables for game success~\cite{Memmert2017}. The development of advanced tracking technologies~\cite{baysal2015sentioscope, liu2009automatic} from a range of companies~\cite{optasports, stats} has opened up new opportunities through the availability of accurate positional data for the players and the ball. These data 
allow to apply automated approaches to analyze different tactical aspects~\cite{Memmert2017, stein2017bring, gyarmati2015automatic}. Referring to \citet{Rein2016}, tactics can be distinguished into team, group and individual tactics. 
One important aspect with respect to team tactics is the team formation~\cite{Bialkowski2014IdentifyingTeamStyle,Rein2016}. The team formation describes the spatial arrangement of the players on the pitch 
by dividing it into tactical groups~(e.g., defenders, midfielders, and attackers). 
%
However, team sports are in general highly complex and dynamic since players are constantly switching positions and roles throughout a match. Consequently, \citet{Bialkowski2014IdentifyingTeamStyle, Bialkowski2016, Bialkowski2014Large-ScaleAnalysis} considered formation detection as a role assignment problem where each player is assigned to the most probable distinct role at each moment of the match at a specific time instant. 
%
%
First automatic approaches for formation classification assumed that the team formation is stable over a match half and thus focus on the classification of formation for whole matches or halftimes~\cite{Bialkowski2014IdentifyingTeamStyle,Bialkowski2014Large-ScaleAnalysis}. More recently, spatio-temporal methods were introduced that aim to detect formation changes during the match~\cite{Bialkowski2016, Machado2017, Wu2019ForVizor}, but either evaluation was not performed on single match situations or, in case of \emph{ForVizor}~\cite{Wu2019ForVizor}, case studies were utilized to evaluate the visual analytics system itself. 
\begin{figure*}[t]
  \centering
  \includegraphics[width=0.92\linewidth]{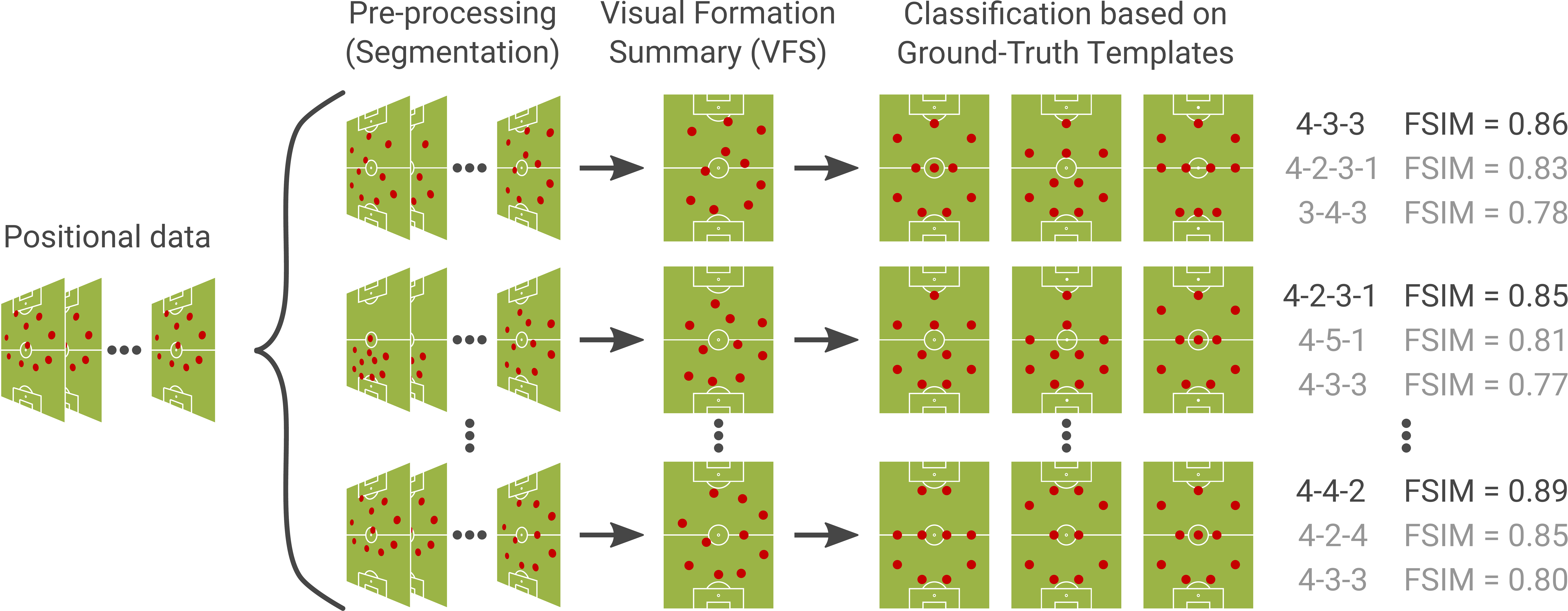}
  \caption{Workflow of the proposed system. The positional data of the a football match are pre-processed~(Section~\ref{sec:input_preprocessing}) to create a series of match situations from ball gain to loss and vice versa. Subsequently, a visual formation summary~(VFS) is generated according to Section~\ref{sec:graph_repr} and compared to a set of ground-truth formation templates for classification~(Section~\ref{sec:classify_formation}).}
  \label{fig:overview}
\end{figure*}

Moreover, previous work~\cite{Bialkowski2016, Wu2019ForVizor} mainly relied on clustering algorithms to automatically find the most prominent formations during a match to measure temporal differences in individual situations. Therefore, it is not possible to directly predict a numerical scheme such as 4-4-2 and the resulting clusters require supervision for classification. In contrast, \citet{Machado2017} developed a visual football match analysis tool where formations are classified by a k-means clustering approach using the coordinates of the players itself and assigning them to one of three tactical groups~(defender, midfielder, attacker). Although this approach is able to directly predict formations, the clustering it is solely based on one-dimensional player coordinates with respect to the field length~(from goal to goal) and formations are restricted to three tactical groups. 

In this paper, we present an analytics system that automatically generates a visualization summary of the formation in single match situations. Thereby, a match situation is defined from gaining to losing ball possession. Based on the visual representation, we present a novel classification approach that calculates the similarity~(of a single match formation) to ground-truth templates for twelve different popular tactical formations~(some examples, e.g., 4-4-2, 4-2-3-1 are shown in Figure~\ref{fig:overview}). The quality of the visualization has been evaluated by twelve domain experts that have also provided ground-truth annotations for numerical schemes of tactical formations. A detailed analysis of the results is provided to evaluate the usefulness of the visualization of formations in single match situations. The inter-coder agreement of the experts is measured and provides first insights about the applicability of numerical schemes to describe team formations. 
%
It turns out that one main issue is that some tactical formations only differ in the interpretation of some roles. To address this issue, we propose a novel metric to measure the quality of the formation classification with respect to the similarity to ground-truth formation templates provided by domain experts. To the best of our knowledge, this is the first work that provides a solution and detailed analysis of formation classification for single match situations. 

The remainder of the work is organized as follows. Section~\ref{chp:rw} reviews related work in sports analytics with focus on formation detection in football games. Our system to create a visual formation summary and to classify the formation in single match situations is presented in Section~\ref{chp:analytics_system}. In Section~\ref{chp:experiments} the experimental results based on expert annotations are discussed in detail. Section~\ref{chp:summary} summarizes the paper and outlines potential areas of future research.
%
\section{Related Work}
\label{chp:rw}
%
Analytics in football or sports in general is a broad field that has recently attracted more attention mostly due to the availability of positional data commonly captured by pre-installed tracking devices in stadiums~\cite{baysal2015sentioscope, liu2009automatic} or provided by companies such as \citet{optasports} or \citet{stats}.  
Since a more general overview goes beyond the scope of this work, we refer to the review of~\citet{Rein2016}, that covers various aspects and challenges of automated content analysis in football. One fundamental research area is the tactical analysis of football data. We therefore focus on related work that has been introduced to find general tactical patterns as well as to explicitly classify and visualize team formations. 

Many approaches have been suggested that aim to cluster and consequently find prominent movement patterns of a team~\cite{Wei2013LargeScaleAnalsysisofFormations,Gudmundsson2014,Wang2015DiscerningTacticalPatterns,Gudmundsson2019}. In this context, sketch-based (video)~retrieval systems were introduced~\cite{Kabary2013SportSenseOriginal,sha2016chalkboarding,Probst2018SportSenseUI} that allow users to draw spatio-temporal queries on a virtual pitch to directly retrieve similar game situations. 
%
While these approaches mainly reveal individual or group tactics, another important factor with significant impact on performance is team formation. However, due to the nature of team sports, players constantly change roles and thus make formation classification a complex task. Based on hockey games, \citet{Lucey2013} have shown that a role-based representation of the formation is superior compared to a representation that is solely based on the coordinates of player identities. Subsequently, \citet{Bialkowski2014Large-ScaleAnalysis,Bialkowski2014IdentifyingTeamStyle} have introduced a role-assignment algorithm and define the formation as a set of role-aligned player positions. However, they assume that the dominant formation is stable within a match half and this coarse temporal granularity is not sufficient to describe the complex and varying tactical formations in modern football.
To solve this issue, Perl et al. presented a number of formation analysis tools~\cite{Grunz2009AnalysisAndSimulationOfActions,Perl2011NetBasedGameAnalysis,Grunz2012TacticalPatternRecognition} and used the neural network~\emph{DyCoN}~\cite{Perl2004DyCon} to determine the distribution and sequential changes in the team formation, respectively. In addition, \citet{Bialkowski2016} have extended their previous systems~\cite{Bialkowski2014Large-ScaleAnalysis,Bialkowski2014IdentifyingTeamStyle} and utilized the role-assignment algorithm to discover with-in match variations using two methods as a proof-of-concept: (1)~clustering of role-aligned player positions and~(2) calculating the distance of each frame to the mean formation of the half time. 
\citet{Wu2019ForVizor} proposed a visual analytics system called \emph{ForVizor}, that distinguishes between offensive and defensive formations. Based on the role-assignment algorithm of~\citet{Bialkowski2014Large-ScaleAnalysis,Bialkowski2014IdentifyingTeamStyle} they subsequently visualize formation changes between different match periods.
%
But the aforementioned systems rely on the detection of the most prominent formations, e.g., by using a clustering algorithm or the average formation of the halftime in order to detect temporal changes in the formation. Therefore, their approach cannot automatically predict a numerical tactical scheme such as 4-4-2 for short match situations.
Alternatively, \citet{Machado2017} developed a match analysis tool 
and applied k-means clustering to the one dimensional $y$~player positions with respect to the field length~(from goal to goal) itself. Each player is then assigned to one of three tactical groups to create a rough but well-known numeric representation. However, this approach completely neglects the $x$-coordinates of the players for classification. 

\section{Team Formation Classification}
\label{chp:analytics_system}
%
As the discussion of related work reveals, previous approaches for team formation analysis focused on entire half-times or matches. In contrast, we present a novel classification approach that can be applied to single match segments of arbitrary length and conduct an in-depth expert evaluation for individual match scenes.
In addition, such an evaluation by domain experts in terms of analyzing individual match situations with respect to the formation played was not conducted yet. Evaluation was either focused on long-term formations~\cite{Bialkowski2014Large-ScaleAnalysis,Bialkowski2014IdentifyingTeamStyle,Bialkowski2016} or placed more emphasis on the evaluation on the tools itself~\cite{Machado2017,Wu2019ForVizor}.

Our proposed system to explore football matches with respect to the team formation is introduced in this section. The definition of a team formation is provided in Section~\ref{sec:definition_formation}. The required input data as well as pre-processing methods are explained in Section~\ref{sec:input_preprocessing}. Based on this input information we propose a methodology to create a visual formation summary~(Section~\ref{sec:graph_repr}) that serves to  classify~(Section~\ref{sec:classify_formation}) the team formation played in single situations of a football match. An overview is illustrated in Figure~\ref{fig:overview}.
%
\subsection{Definition of Team Formation}
\label{sec:definition_formation}
%
In general, the team formation describes the spatial arrangement of players within a team. Assuming that all ten players~(except the goalkeeper) are on the pitch, it is defined as a set of ten distinct roles~$F=\{r_1, \ldots, r_{10}\}$ that are represented by their two-dimensional position~$r \in \mathbb{R}^2$ on the football field. For simplification, these roles are often assigned to tactical groups like defenders, midfielders (defense and offensive) and attackers in order to generate a numeric representation. 
These numerical schemes, e.g. 4-2-3-1 and 4-4-2, define the tactical formation of the team and are denoted as~$F_n$ in the following.
%
\subsection{Input Data and Pre-processing}
\label{sec:input_preprocessing}
%
Our system relies on two-dimensional location information of each player at each discrete timeframe~$f$. We use the normalized coordinates with respect to the width~$x \in [0.0, 0.7]$ and length~\mbox{$y \in [0.0, 1.0]$} of the football pitch and preserve the aspect ratio of the field. For unification, the direction of play of the observed team is always considered from bottom to top and the position data of the players are converted accordingly. 
Since the aim of this work is to automatically detect formations in individual game situation, the match is first divided into temporal segments. 
In this context, we require information which team possesses the ball at each timeframe~$f$ and define a game segment~$S=\{f_i, \dots , f_{i+m}\}$ containing $m$~timeframes from gaining to losing the ball or vice versa.
%
\subsection{Visual Formation Summary}
\label{sec:graph_repr}
%
To classify the formation, first a \emph{visual formation summary}~(VFS) from the two-dimensional position data of all frames in a game segment~$S$ is generated. Regardless of how the players of the observed team have moved on the pitch, e.g., while defending most players are located in the own half, the formation in terms of the distance between the players within a team remains the same. Therefore, we subtract the team center that is defined as the mean of all individual player positions at each timeframe for normalization. 
As stated in Section~\ref{sec:definition_formation} the formation is defined by a set of roles. Theoretically, each player can be considered to act in one role and represented by its mean position during a match segment. However, as mentioned by previous work~\cite{Bialkowski2014Large-ScaleAnalysis, Bialkowski2016}, players can potentially switch roles and this approach would not accurately reflect the tactical formation. For this reason, we employ the role-assignment algorithm of \citet{Bialkowski2014Large-ScaleAnalysis, Bialkowski2016} to detect and compensate role changes with the change that only one iteration is applied. More than one iteration did not have a great influence on the result in our experiments, which is supposedly due to the length of the game sequences. 
Finally, the mean position~$\overline{r}$ for each role during the observed match segment is utilized to define the formation~$F = \{\overline{r}_1, \dots, \overline{r}_{10}\}$ and to derive the visual formation summary.
%
\subsection{Classification of Numerical Schemes}
\label{sec:classify_formation}
%
The formation~$F$ with compensated role changes according to the previous section serves as input to classify each game situation into a common numerical tactical schema like 4-4-2. We propose a novel classification approach that measures the similarity of the extracted formation~$F$ to a pre-defined set of $t$~popular football \mbox{formations~$T=\{\hat{F}_1, \dotsc, \hat{F}_t\}$}. The expected player coordinates 
are provided by domain experts as explained in Section~\ref{sec:templates}. 

In order to enable a comparison between two formations, it is necessary to normalize the positional data of each role~$\tilde{r}$ by the minimum and maximum $x$ and $y$~coordinate within a formation~$F$:
\begin{equation}
    \tilde{r} = \frac{\overline{r} - \min(F)}{\max(F) - \min(F)} \qquad \forall \, \overline{r} \in F \, .
\end{equation}
The formula provides also a normalization of the relative distances of the players and therefore allows for a comparison of formations with different compactness.
Subsequently, a similarity matrix~\mbox{$M(F_1, F_2) \in \mathbb{R}^{10\times10}$} is calculated. Since we use idealized templates for formation classification, the euclidean squared distance is applied in this context, because it penalizes smaller distances between different roles less severely. Each entry~$m_{i,j}$ is then calculated based on the positional coordinates of each role~$~\tilde{r}_i = (\tilde{x}_i, \tilde{y}_i)$ in formation~$F_1$ to each role~$\tilde{r}_j = (\tilde{x}_j, \tilde{y}_j)$ in formation~$F_2$ according to the following equation:
\begin{equation}
\label{eq:sim_matrix}
    M(F_1, F_2) = \max{\left(1 - \frac{||\tilde{r}_i - \tilde{r}_j||_2^2}{\delta}, 0\right)} \qquad \forall \, \tilde{r}_i \in F_1; \tilde{r}_j, \in F_2 \, .
\end{equation}
The normalization factor~$\delta$ serves as tolerance radius. Under the assumption that a football pitch can be divided into three horizontal~(left, center, right) and vertical groups~(defender, midfielder, attacker), a normalization factor~$\delta = \nicefrac{1}{3}$ means that the similarity of wingers to central player as well as, e.g., from attackers to midfielders would already become zero. In our opinion, this fits well to the task of formation classification. Please note, that we only allow similarity values in the interval range~$m(i,j) \in [0, 1]$. 

To calculate the similarity of two formations, each role in formation~$F_1$ has to be assigned to its optimal counterpart in formation~$F_2$. 
With the constraint that each role can only be assigned once and the overall goal to maximize the similarity 
this results in a linear sum assignment problem that can be solved via the \emph{Hungarian algorithm}~\cite{Kuhn1955Ass} whose solution corresponds to: 
%
\begin{equation}
\label{eq:hungarian}
m^*_{i, j} = \begin{cases}
    m_{i,j}     & \text{, if } \tilde{r}_i \in F_1 \text{ is assigned to } \tilde{r}_j \in F_2, \\
    0           & \text{otherwise.}
\end{cases}
\end{equation}
Finally, the formation similarity~FSIM($F_1, F_2$) of the compared formations is defined as the sum of all elements in the similarity matrix~$M^*(F_1, F_2)$ normalized with respect to the number of assigned roles~(in this case ten).
%
%
To classify the derived formation~$F$ according to Section~\ref{sec:graph_repr} of a given match segment into the numerical schema~$F_n$, we calculate the similarities to a set of pre-defined templates~$T=\{\hat{F}_1, \dotsc, \hat{F}_t\}$ that contain idealized role positions for $t$~popular football formations. This allows us to generate a ranking of the most probable numerical formation played in an individual match situation. For the final classification, the template formation with the highest similarity is selected as defined in Equation~\eqref{eq:codebook_selection}:
\begin{equation}
\label{eq:codebook_selection}
    F_n^*= \underset{ \hat{F} \in T}{\argmax}\left(\text{FSIM}(F,\hat{F})\right) \,
\end{equation}
%
%
%
\section{Experimental Results}
\label{chp:experiments}
%
In this section, the experimental setup with details on the dataset characteristics and the annotation study with domain experts are presented~(Section~\ref{sec:setup}). Furthermore, the templates with idealized role position for twelve popular football formation are presented in Section~\ref{sec:templates}. The evaluation metrics including a novel similarity measurement based on ground truth formations are introduced in Section~\ref{sec:metrics}. In-depth analysis of the quality of the ground-truth annotations as well as the evaluation of the visual formation summary~(VFS) are presented in Section~\ref{sec:expert_study} and Section~\ref{sec:eval_graphic}. Finally, the results of the automatic formation classification to reveal team tactics are discussed in Section~\ref{sec:eval_cls}.
\subsection{Setup}
\label{sec:setup}
\subsubsection{Dataset:}
The dataset contains four matches that took place in the 2011/2012 \emph{first league season} (omitted for double blind review). The positional data were captured by a \emph{VisTrack} device with a temporal resolution of $25$~frames per second, that  
provides further information like events~(corners, free kicks, etc.), actions~(passes, shots, etc.), ball possession and game status~(running or interrupted). 
We use the additional events and game status to further clean the data as explained in the following. 
%
First, we only consider frames with game status \emph{running} to get rid of all interruptions. Based on the information about ball possession, the matches are temporally segmented according to Section~\ref{sec:input_preprocessing}. In this context, we discard all match segments that a shorter than five seconds since they most likely do not contain any valuable tactical information. Since standard situations resolve tactical formations, we remove all frames five seconds after throw-ins and ten seconds after free kicks, corners, and penalties. The recognition of tactical patterns in this kind of situations should be analyzed independently since they show different characteristics, which is beyond the scope of the current work. 
For a more detailed analysis we furthermore distinguish between short~($5\,s \leq t < 10\,s$), medium~($10\,s \leq t < 20\,s$) and long match situations~($t \geq 20\,s$) and consider situations of own and opposing team's ball possession independently in the experiments.
\subsubsection{Annotation Study:}
\begin{table}[t]
\caption{Number of annotations, mean match situation length and standard deviation~($\sigma$) of the dataset used for evaluation. 
A subset of match situations was annotated by two domain experts each in order to assess the quality of the annotations.} 
\label{tab:dataset_stats}
\begin{tabular}{c | c | c c}
Ball        & \multirow{2}{*}{Duration}     & Annotations           & Mean length [s]   \\
Possession  &                               & (from two experts)    & $\sigma$ [s]      \\
\hline \hline
\multirow{3}{*}{Own}        & short & 105 (35)  & 6.96 (1.38)   \\
                            & mid   & 234 (74)  & 13.67 (2.87)  \\
                            & long  & 231 (71)  & 31.53 (12.62) \\
\hline  
\multirow{3}{*}{Opponent}   & short & 91 (23)   & 6.73 (1.30)   \\
                            & mid   & 232 (71)  & 13.71 (2.95)  \\
                            & long  & 236 (74)  & 32.17 (13.35) \\

 
 
\end{tabular}
\end{table}

The annotation study was split into two parts to (1)~provide ground-truth annotations of the tactical formation for a given match situation and (2)~evaluate the respective visual formation summary~(VFS). Twelve domain experts~(professional game analysts) from an \emph{Institute of Sports Sciences}~(omitted for double blind review) were available in both parts, who had a total time of 60~minutes to annotate 100~situations. The experts received two sets that each contained $50$~scenes that were evenly sampled from a single half of a football match. The sets contained $25$~scenes for both own and opposing team's ball possession. The scenes were watched in chronological order to simulate a real analysts process and allow the annotator to benefit from contextual knowledge of previous situations. To assess the quality of the annotations given by the experts, eight sets of 50~match situation~(total of 400 annotations) were assigned to two experts each. Detailed statistics are presented in Table~\ref{tab:dataset_stats}. Note that $71$~formation annotations and $307$~ratings for the VFS are missing due to the time constraint.

\begin{figure}[t]
  \centering
  \includegraphics[width=1.00\linewidth]{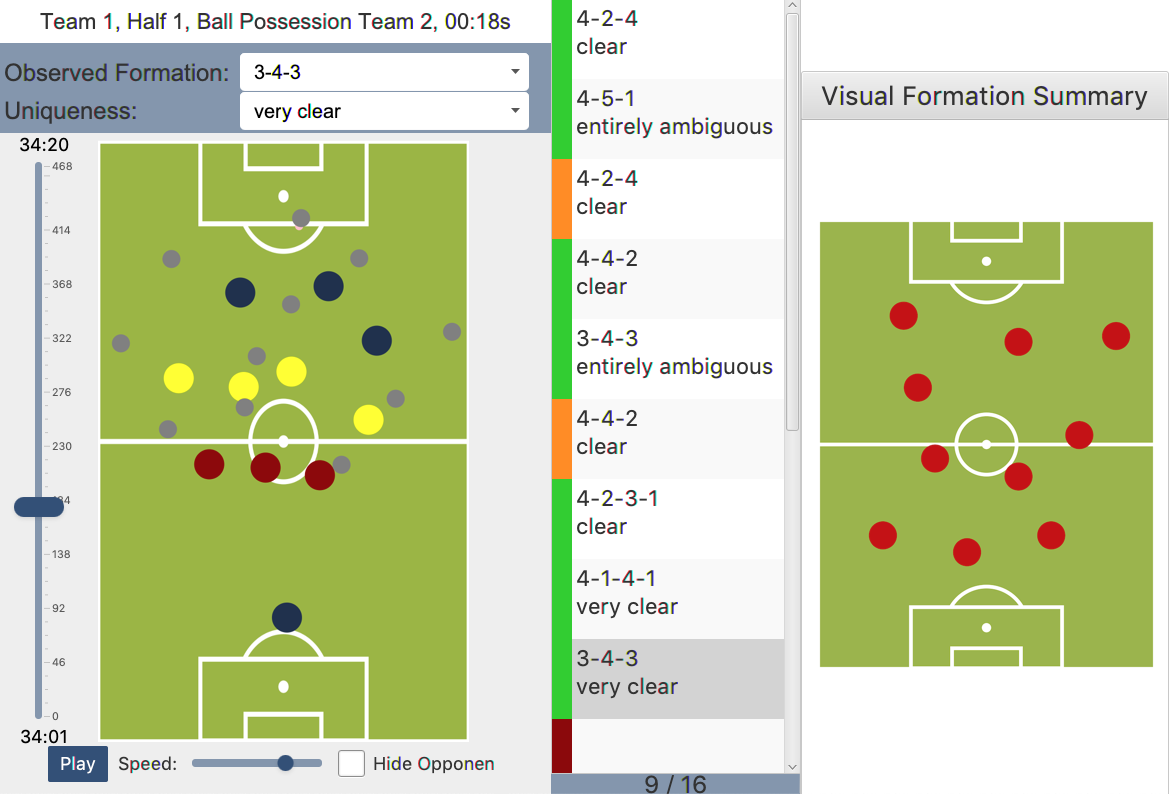}
  \caption{Analytics tool for formation detection. The two-dimensional animation of the selected match situation~(middle) is shown on the left and the resulting visual formation summary of the scene is shown on the right side.}
  \label{fig:gui}
\end{figure}

\begin{figure}[t]
  \centering
      \includegraphics[width=1.00\linewidth,trim={0.0cm 0.00cm 0.2cm 0.0cm},clip]{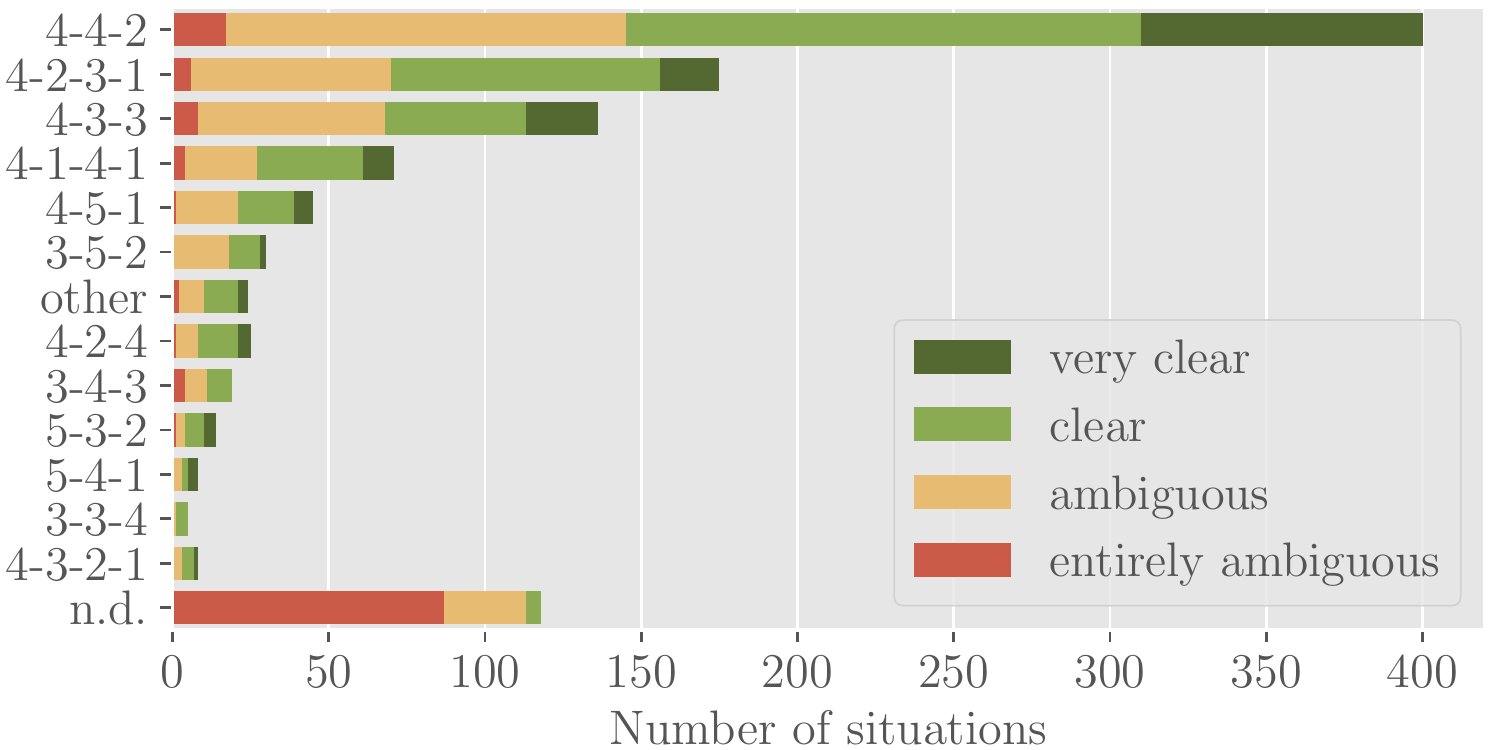}
  \caption{Distribution of annotated tactical formations and the degree of clarity~(emphasized in different colors).}
  \label{fig:annotation_stats}
\end{figure}

\textbf{Ground-truth Formation Annotations:} At first the annotator was asked to watch the two-dimensional match animation based on positional data of all players~(players of the opposing team were displayed opaque for reference) and the ball of a given match segment. The situation could either be explored by playing the scene automatically in real-time or manually using a slider. To help finding tactical groups, specific players could be marked by different colors. The analytics tool is shown in Figure~\ref{fig:gui}. 
Finally, the annotator was able to select one of twelve different formations as well as the options \emph{other} and \emph{undefined}~(as listed in Figure~\ref{fig:annotation_stats}) if a formation cannot be assigned unambiguously. Besides the annotation of the tactical scheme such as 4-4-2, the annotator had to rate how well the formation could be determined 
by choosing one of the following options: \emph{entirely ambiguous, ambiguous, clear, very clear}. The statistics of annotated match situations is illustrated in Figure~\ref{fig:annotation_stats}.

\textbf{Rating of the Visual Formation Summary:} After providing the ground-truth formation, the annotator had to evaluate the VFS of the observed scene. In this context, he was allowed to select one of the three following options: \emph{bad, neutral, good}.
%
%

\begin{table*}[t]
\setlength{\tabcolsep}{6pt} 
\caption{Agreement of the expert annotations for detecting a formation in terms of the agreement accuracy, formation similarity~(FSIM) and Krippendorff's $\alpha$. Top-k accuracy means that the specified formation of at least one annotator is within the top-k nearest formations~(according to Figure~\ref{fig:template_similarities}) of the specified formation of the other annotator.} 
\label{tab:study-agree}
\begin{tabular}{c|c|cccc|cccc|c}

\multicolumn{2}{c|}{Ball Possession} & \multicolumn{4}{c|}{Own} & \multicolumn{4}{c|}{Opponent} & \multicolumn{1}{c}{\multirow{2}{*}{Overall (207)}} \\ \cline{1-10}
\multicolumn{2}{c|}{Duration (Nr. of scenes)} & \multicolumn{1}{c}{short (13)} & \multicolumn{1}{c}{mid (39)} & \multicolumn{1}{c}{long (43)} & \multicolumn{1}{c|}{all (95)} & \multicolumn{1}{c}{short (10)} & \multicolumn{1}{c}{mid (44)} & \multicolumn{1}{c}{long (58)} & \multicolumn{1}{c|}{all (112)} & \multicolumn{1}{c}{} \\ \hline \hline
\multirow{3}{*}{Accuracy} 
& Top-1 & 0.38 & 0.41 & 0.42 & 0.41 &   0.70 & 0.55 & 0.55 & 0.56 &     0.49 \\
& Top-3 & 0.62 & 0.69 & 0.70 & 0.68 &   0.80 & 0.70 & 0.76 & 0.72 &     0.71 \\
& Top-5 & 0.77 & 0.82 & 0.95 & 0.87 &   1.00 & 0.95 & 0.98 & 0.97 &     0.93 \\ \hline
\multicolumn{2}{c|}{FSIM} & 0.90 & 0.91 & 0.92 & 0.91 &     0.96 & 0.94 & 0.94 & 0.94 & 0.93 \\
\multicolumn{2}{c|}{Krippendorff's $\alpha$} & 0.23 & 0.25 & 0.18 & 0.22 &  0.54 & 0.20 & 0.26 & 0.27 & 0.26 \\

\end{tabular}
\end{table*}
\subsection{Template Formations}
\label{sec:templates}
\begin{figure}[t]
  \centering
  \includegraphics[width=1.0\linewidth]{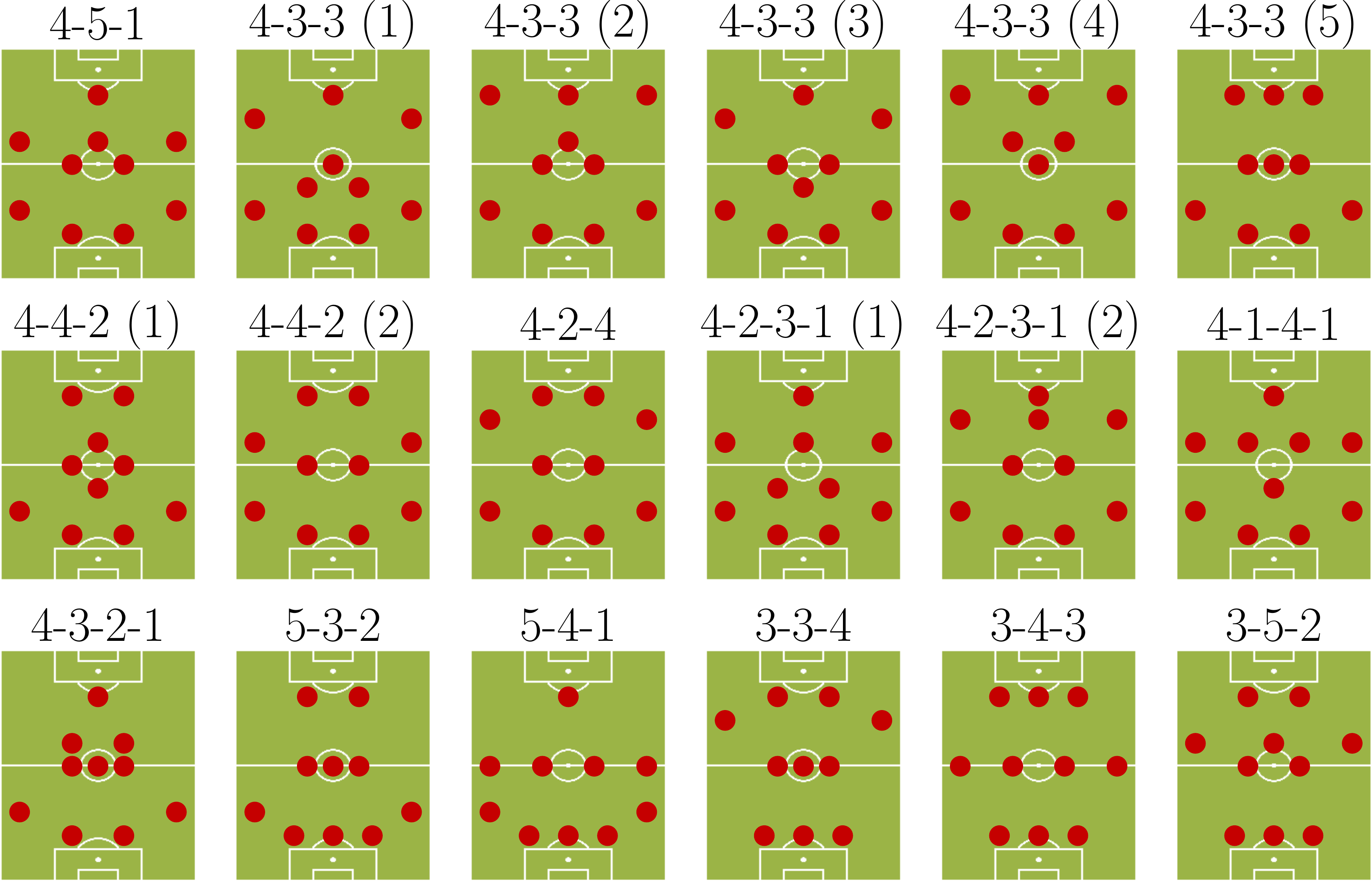}
  \caption{Templates with idealized role positions for twelve popular football formations created by domain experts. If a formation is ambiguous, several variations are provided.}
  \label{fig:templates}
\end{figure}
As explained in Section~\ref{sec:classify_formation}, the classification of the played formation is performed by a comparison to a pre-defined set of templates for different tactical formations. 
Our domain experts were asked to provide these ground-truth templates to the best of their knowledge. But some formations like 4-4-2 contain some variations and are not completely unambiguous. Hence, multiple templates for one formation should be created. For classification we have calculated the similarity of the visual formation summary to all variations of a single formation, and used the maximum as value for the formation similarity~FSIM. The templates created for all twelve formations used in the experiments are visualized in Figure~\ref{fig:templates}.
\subsection{Evaluation Metrics}
\label{sec:metrics}
In order to assess the quality of the visual formation summary~(VFS) of a given match sequence, we propose to calculate the formation similarity~FSIM to the template of the annotated ground-truth formation as an evaluation metric. The accuracy is calculated to measure the classification performance of the system. Since the data set contains a large bias towards some formations, we report micro-accuracy alongside macro-accuracy as it allows us to study system performance while considering each class to be equally important. However, as already mentioned, some formation such as 3-5-2 and 5-3-2 show very similar spatial characteristics and only depend on the subjective interpretation of some player roles. Hence, the VFS is compared to all available template formations. This allows us to generate a ranking with respect to the formation similarities~FSIM and additionally report the top-$k$ accuracy. Please note, that some match situations were analyzed by two experts and their annotations can differ. But we assume that both annotations are valid and use the annotated formation which has a higher similarity to the VFS as reference.

%

\subsection{Analysis of the Expert Study}
\label{sec:expert_study}
In our experiments, we only consider match situations where a formation was \emph{clearly} or \emph{very clearly} recognizable for at least one expert. This results in a total of 472 unique situations of which 207 were annotated by two experts for classification and 450 ratings for the visual formation summary.
Referring to Figure~\ref{fig:annotation_stats}, the analysis of the annotated match situations has shown a huge bias towards some popular formations such as 4-4-2, 4-2-3-1 and 4-3-3. These results were more or less expected, since e.g. the 4-4-2 is generally widely accepted and therefore used more frequently to describe a formation compared to a 4-2-4, which however has very similar spatial properties as shown in Figure~\ref{fig:templates}. More surprisingly, the majority of annotations were rated at least \emph{clearly} recognizable by the experts, despite the short length of single match situations. In this context, the defensive formations were annotated with a larger confidence than offensive formations and have shown less variance~(mainly 4-4-2). This effect can be explained by the increased freedom of the players during the attack to make creative plays, which are very important for scoring goals in modern football~\cite{kempe2018good}. 

\begin{figure}[t]
  \centering
  \includegraphics[width=1.0\linewidth,trim={0.25cm 0.45cm 0.1cm 0.4cm},clip]{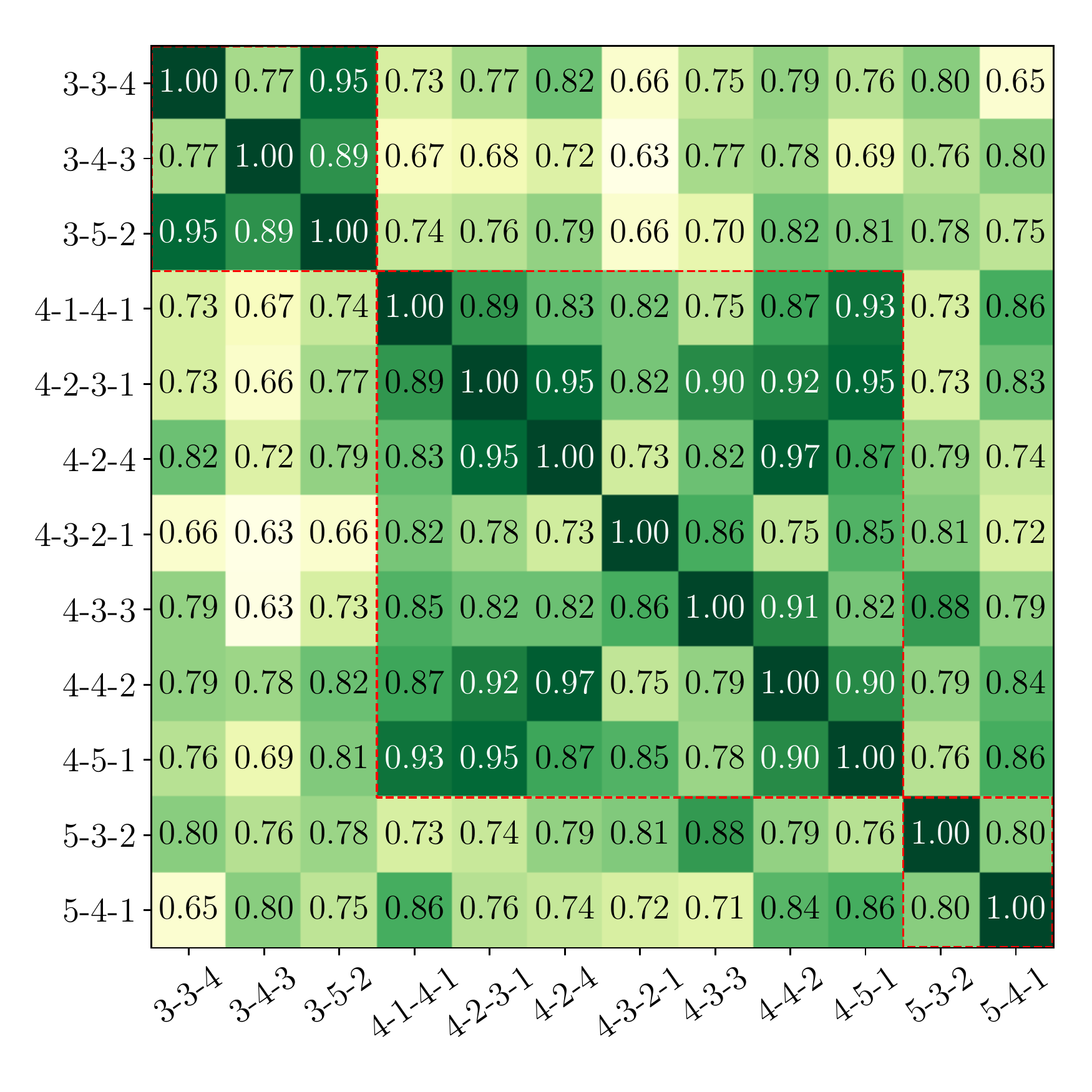}
  \caption{Formation similarities~(FSIM) of the templates with idealized player positions~(Figure~\ref{fig:templates}) of twelve popular football formations. Formations with the same amount of defenders show very high similarities~(emphasized with red borders).}
  \label{fig:template_similarities}
\end{figure}

In order to assess the quality of the provided annotation of the played formations, the inter-coder agreement is measured. In this context, we only consider match situations in which at least one expert has clearly or very clearly identified the formation. The results are reported in Table~\ref{tab:study-agree}. In particular, the agreement in terms of \emph{Krippendorff's}~$\alpha$ and the top-1 accuracy is noticeably lower than expected. Overall, annotations for defensive formations show significantly more correlation than the annotated offensive formations. As already mentioned, this is mainly due to freedom and creativity in attacking situations that lead to more fluid formations. 
However, we believe that the annotations from domain experts still show correlations, but that the complexity and subjectivity of the task leads to different conclusions. As stated above, team formations sometimes only differ in the interpretation of very specific roles. In addition, it is possible that (1)~formations are not symmetric and (2)~different formations are played within a single game situation, e.g. when multiple offensive game patterns are performed during an attack. Therefore, we propose to calculate the formation similarity~FSIM between the templates of the annotated formations to obtain an alternative measure of the inter-coder agreement. This also enables us to determine the top-k accuracy since a ranking of the most similar formations can be generated. The similarity values of all tactical formations are visualized in Figure~\ref{fig:template_similarities}. 
%
It is clearly visible that the top-3 accuracy is significantly better than the top-1 accuracy~(Table~\ref{tab:study-agree}) with respect to the inter-coder agreement. In addition, the formation similarities~FSIM between the annotated formations are comparably high, especially if the values from Figure~\ref{fig:template_similarities} are taken into account. From our point of view, this indicates that the annotations of experts do indeed show a high correlation, at least if the recognition of formations is treated as a multi-label task where more than one answer can be considered as correct. 
\subsection{Evaluation of the VFS}
\label{sec:eval_graphic}
In the second part of the annotation study, the domain experts were asked to rate the usefulness of the extracted visual formation summary~(VFS) of a team in a given match situation. In addition, we quantified the similarity of the VFS to the template of the annotated formation. The results are reported in Table~\ref{tab:study-eval-graphic}. 
\begin{table}[t]
\setlength{\tabcolsep}{5pt}
\caption{Evaluation results of the graphical representation created according to Section~\ref{sec:graph_repr} in terms of expert ratings and the formation similarity~FSIM to the annotated formation of the expert.} 
\label{tab:study-eval-graphic}

\begin{tabular}{c|c|ccc|c} 

\textbf{Ball}   & \textbf{Duration} & \multicolumn{3}{c|}{\textbf{Rating}}  & \textbf{FSIM} \\
\textbf{possession} & \textbf{(\#Scenes)}   & Bad   & Neutral   & Good             \\
\hline
\hline
\multirow{4}{*}{Own}        & short (23)    & 0.26  & 0.30      & 0.43       & 0.77 \\
                            & mid (87)      & 0.22  & 0.33      & 0.45       & 0.76 \\
                            & long (96)     & 0.15  & 0.38      & 0.48       & 0.76 \\
                            & all (206)     & 0.19  & 0.34      & 0.46       & 0.76 \\
\hline 
\multirow{4}{*}{Opponent}   & short (24))   & 0.08  & 0.33      & 0.58       & 0.78 \\
                            & mid (98)      & 0.14  & 0.26      & 0.60       & 0.80 \\
                            & long (122)    & 0.16  & 0.30      & 0.55       & 0.81 \\
                            & all (244)     & 0.14  & 0.28      & 0.57       & 0.80 \\
\hline
\multicolumn{2}{c|}{Overall (450)}          & 0.16  & 0.31      & 0.52       & 0.78 \\

\end{tabular}
\end{table}
Overall, the VFS's were mainly rated positive and only in $16\%$ of the situations the annotator did not see correlations to the two-dimensional schematic visual representations. This demonstrates that the VFS indeed provides a good overview in the majority of the cases. Particularly, in situations where the opponent of the observed team possesses the ball, the VFS can quickly give insights into the tactical defensive formation and therefore simplifies the analysts' process.
The same conclusions can be drawn with respect to the obtained formation similarity of the extracted VFS to the templates of the annotated formation. Similarities around 0.75 and 0.80 are achieved in the two cases of own and opposing team ball possession, respectively. Although these values are comparatively lower than the template similarities in Table~\ref{fig:template_similarities}, we believe that the results indicate a satisfying system quality. The template similarities are calculated based on idealized role positions that can be partially shared between two templates. Therefore, the values are expected to be higher, since the VFS of real football situations show more variations.
\subsection{Evaluation of the Formation Classification}
\label{sec:eval_cls}
\quad\textbf{Baselines:} As discussed in the related work section, previous work~\cite{Bialkowski2016, Wu2019ForVizor} apply clustering approaches to find the most prominent formations in a match in order to measure formation changes. These approaches are not capable to automatically classify the formation 
and are therefore not suitable for comparisons. For this reason, we can only compare our proposed classification approach to \citet{Machado2017}'s system. However, their solution relies on a k-means clustering of y-coordinates and can only predict a pre-defined amount of, in this case $k=3$, tactical groups. In addition, it could predict unrealistic formations such as 2-7-1. This is a systematic drawback and the expert annotations shown in Figure~\ref{fig:annotation_stats} indicate that \emph{other} formations are labelled very rarely. 

\textbf{Comparison to baseline approaches:} To enable a comparison, we reduce the number of groups in the annotated formations as well as predictions from four to three by assigning the most similar formation with three tactical groups according to Figure~\ref{fig:template_similarities}. The annotated 4-1-4-1 and 4-2-3-1 formations become a 4-5-1 and the 4-3-2-1 is converted to a 4-3-3, yielding a total number of nine different classes. \citet{Machado2017}'s classification approach based on the clustering of y-coordinates is also applied to the visual formation summaries and thus to the same input data as our system. In addition, we investigate the impact of the role assignment algorithm. The results are reported in Table~\ref{tab:eval-sota}.
\begin{table}[t]
\setlength{\tabcolsep}{6pt}
\caption{Classification results for maximum three and four tactical groups~$|G|$ of the baseline approach from \citet{Machado2017} as well as our proposed system with and without the role compensation algorithm~\cite{Bialkowski2014Large-ScaleAnalysis}. Results are reported for all 472 scenes where at least one annotator could clearly or very clearly identify a formation.} 
\label{tab:eval-sota}

\begin{tabular}{c l |cc|cc} 


\multirow{2}{*}{$|G|$}  & \multirow{2}{*}{\textbf{Method}}   & \multicolumn{2}{c|}{\textbf{Macro-ACC}} & \multicolumn{2}{c}{\textbf{Micro-ACC}} \\
                        &                           & Top-1     & Top-3         & Top-1     & Top-3 \\
\hline
\hline
\multirow{5}{*}{3}      & Random guess              & 0.11      & 0.33          & 0.11      & 0.33  \\
                        & Always 4-4-2              & 0.43      & ---          & 0.11      & ---  \\
                        & K-means~\cite{Machado2017} & 0.14      & ---           & 0.14      & ---  \\
                        & Ours without RC           & 0.15      & 0.40          & 0.18      & 0.38  \\
                        & Ours with RC & \textbf{0.22} & \textbf{0.61} & \textbf{0.24} & \textbf{0.48}  \\
\hline
\multirow{4}{*}{4}      & Random guess              & 0.08      & 0.25          & 0.08      & 0.25  \\
                        & Always 4-4-2              & 0.43      & ---          & 0.08      & ---  \\
                        & Our without RC            & 0.19      & 0.49          & 0.17      & 0.37  \\
                        & Ours with RC & \textbf{0.20} & \textbf{0.52} & \textbf{0.20} & \textbf{0.39}  \\
\end{tabular}
\end{table}
The results clearly show that our classification approach is superior to \citet{Machado2017}'s approach. Furthermore, we could confirm that the role assignment algorithm improves team formation classification.

\begin{figure*}[t]
  \centering
  \parbox{.61\linewidth}{
    \includegraphics[width=1.00\linewidth]{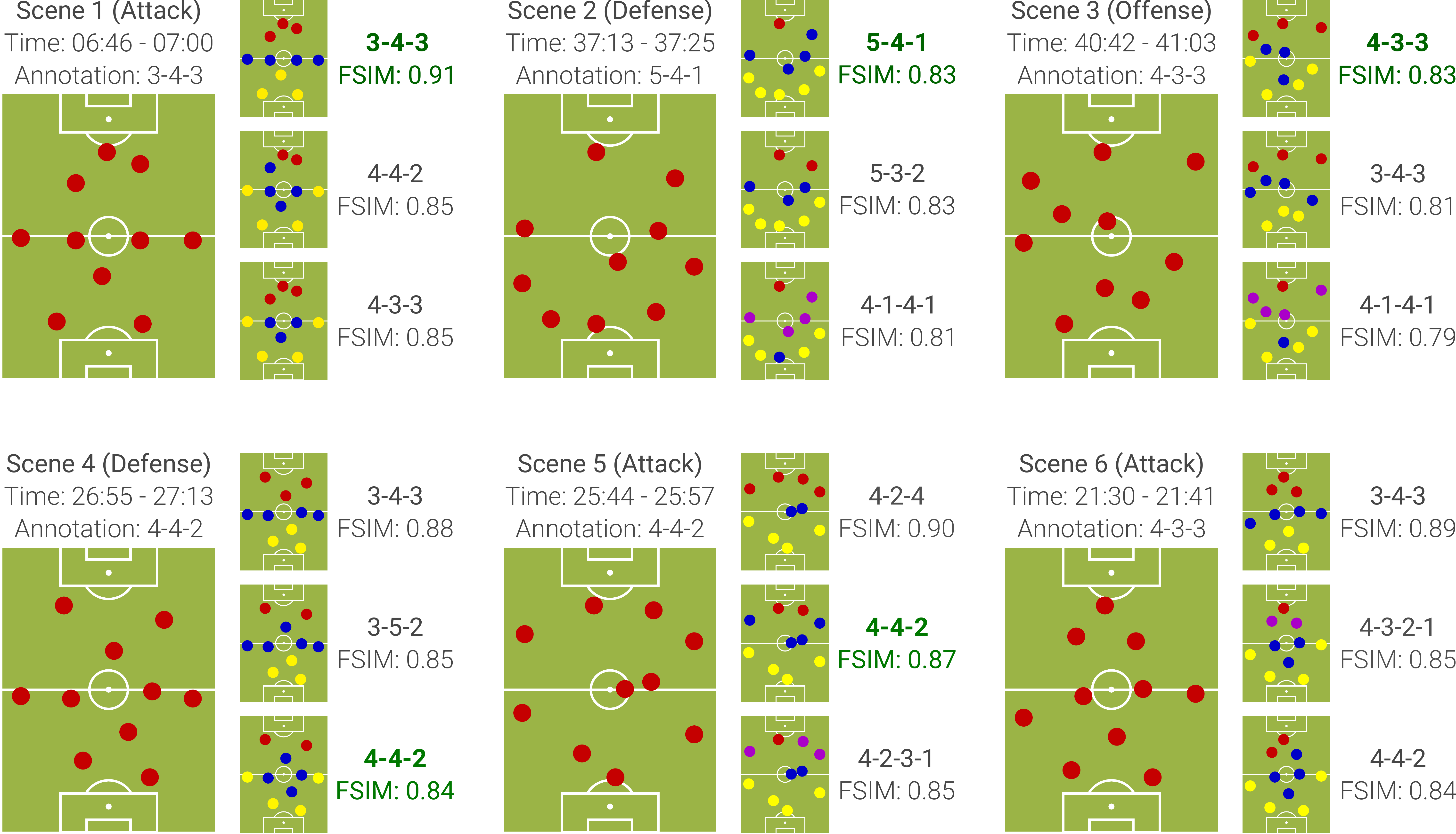}
  }
  \hfill
  \parbox{.37\linewidth}{
    \includegraphics[width=1.0\linewidth,trim={0.25cm 0.45cm 0.1cm 0.4cm},clip]{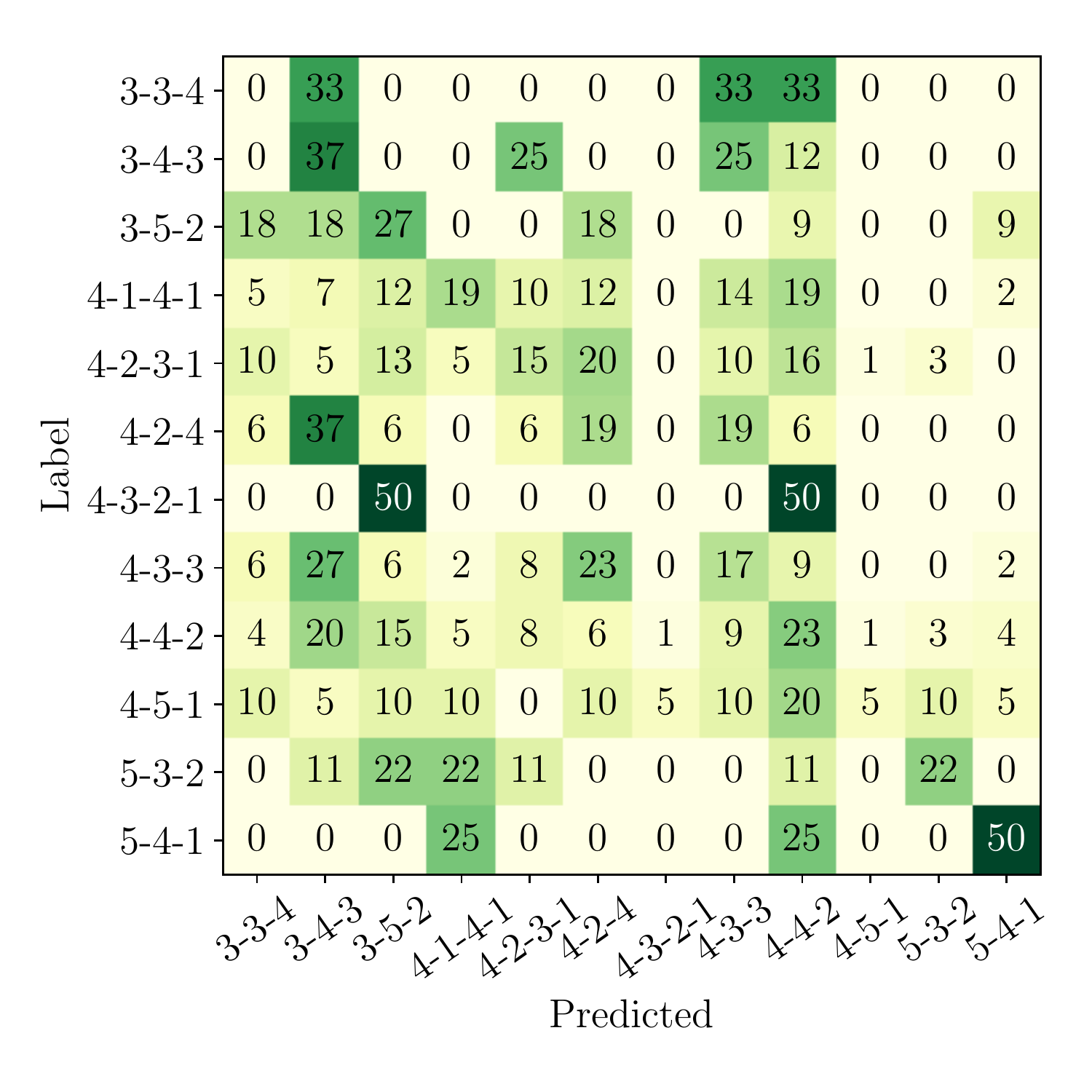}
  }
  \caption{Qualitative results~(left) of the proposed analytics system for six individual scenes with the respective top-3 tactical group assignment~(emphasized in different colors) after comparing with the templates in Figure~\ref{fig:templates} as well as the confusion matrix~(right) for the predictions of all 472 match situations in percent.}
  \label{fig:qual_results}
\end{figure*}

\textbf{Analysis and discussion of the results:} Although the results are improved compared to previous baselines, in particular the top-1 accuracy is rather low. To analyze possible problems in more detail, quantitative as well as qualitative results are provided in Table~\ref{tab:eval_cls} and Figure~\ref{fig:qual_results}, respectively.  

\begin{table*}[t]
\setlength{\tabcolsep}{5pt} 
\caption{Evaluation results of the classification approach in terms of the micro and macro top-k accuracy. Note, that only match situations are considered in which at least one expert could clearly or very
clearly identify the formation.}
\label{tab:eval_cls}
\begin{tabular}{c|c|cccc|cccc|c}

\multicolumn{2}{c|}{Ball Possession} & \multicolumn{4}{c|}{Own} & \multicolumn{4}{c|}{Opponent} & \multicolumn{1}{c}{\multirow{2}{*}{Overall (472)}} \\ \cline{1-10}
\multicolumn{2}{c|}{Duration (Nr. of scenes)} & \multicolumn{1}{c}{short (31)} & \multicolumn{1}{c}{mid (94)} & \multicolumn{1}{c}{long (97)} & \multicolumn{1}{c|}{all (222)} & \multicolumn{1}{c}{short (30)} & \multicolumn{1}{c}{mid (100)} & \multicolumn{1}{c}{long (120)} & \multicolumn{1}{c|}{all (250)} & \multicolumn{1}{c}{} \\ \hline \hline
\multirow{5}{*}{Macro Accuracy} & Top-1 & 0.13 & 0.09 & 0.13 & 0.11 &   0.30 & 0.24 & 0.30 & 0.28 & 0.20 \\
& Top-2 & 0.19 & 0.20 & 0.29 & 0.24 &   0.40 & 0.47 & 0.45 & 0.45 & 0.35 \\
& Top-3 & 0.45 & 0.33 & 0.52 & 0.43 &   0.53 & 0.63 & 0.60 & 0.60 & 0.52 \\
& Top-4 & 0.55 & 0.51 & 0.66 & 0.58 &   0.77 & 0.71 & 0.73 & 0.73 & 0.66 \\
& Top-5 & 0.61 & 0.63 & 0.73 & 0.67 &    0.80 & 0.80 & 0.81 & 0.80 & 0.74 \\ \hline
\multirow{5}{*}{Micro Accuracy} & Top-1 & 0.20 & 0.09 & 0.15 & 0.10 &   0.19 & 0.28 & 0.29 & 0.23 & 0.20 \\
& Top-2 & 0.24 & 0.16 & 0.25 & 0.16 &   0.25 & 0.38 & 0.36 & 0.30 & 0.27 \\
& Top-3 & 0.43 & 0.23 & 0.40 & 0.25 &   0.33 & 0.58 & 0.46 & 0.40 & 0.39 \\
& Top-4 & 0.46 & 0.34 & 0.49 & 0.33 &   0.58 & 0.61 & 0.53 & 0.53 & 0.48 \\
& Top-5 & 0.49 & 0.48 & 0.57 & 0.46 &   0.63 & 0.64 & 0.58 & 0.56 & 0.53 \\ \hline
\end{tabular}
\end{table*}
The quantitative results again show that the task of formation classification becomes easier for defense situations and for situations with increasing length. The results for the setting \emph{all situations} are much better than random guessing and is also improved significantly when considering the top-k, particularly for $k>2$, similar formations. Referring to the qualitative results in Figure~\ref{fig:qual_results}, this can be mainly explained by the problems described below. 
As previously stated, formations can be very similar and their classification often depends on the interpretation of very specific roles. In particular, wing backs in formations with four defenders are moving up on the pitch to get involved in the attack or pro-actively defend in pressing situations. This is clearly visible in scenes~1, 4 and 6. While the domain expert in scene~1 considers them as midfielders, they are mostly perceived as defenders in similar situations. 
However, simultaneously the defensive midfielder could move back to form a three-man formation with both center backs. For this reason, e.g., a 4-4-2 or 4-3-3 is often classified as a 3-4-3 or 3-5-2 by our system, as depicted in the confusion matrix~(Figure~\ref{fig:qual_results} right). Similarly, offensive wingers or attacking midfielders can be either interpreted as midfielders or strikers. Referring to Figure~\ref{fig:annotation_stats}, the experts lean towards more popular formations such as 4-4-2 during their annotations, while the visual formation summary suggests a 4-2-4 instead with respect to the pre-defined templates~(scene~5). 
Admittedly, the experts have classified the formation based on the two-dimensional graphical animation representation and in addition had context from preceding situations which can have influence on the rating and allows for other conclusions. But in many cases the annotators even considered the VFS as \emph{good}, which shows that the mistakes are often connected to the subjective interpretation of roles instead of the similarity to idealized templates.
Overall, the analysis suggests that formation classification should be considered as a multi-label or fuzzy classification task, where more than one answer could be correct. For this reason, we believe that a visual formation summary often provide valuable insights into the tactical formations. The formations' similarities to templates of popular formation, however, can help to monitor tactical changes as well as to retrieve situations to show specific formations. 

%
\section{Conclusions and Future Work}
\label{chp:summary}
%
In this paper, we have presented an analytics approach for the visualization and classification of tactical formations in single situations of football matches. A novel classification approach has been proposed employing a set of ground truth templates that contains idealized player positions for twelve popular team formations. 
A detailed analysis of an expert annotation study was conducted to provide results for defensive and offensive formation classification in match situations with various length. The study has clearly demonstrated the complexity of the task, particularly for offensive formations, since even annotations from domain experts differ due to the subjectivity in interpreting roles of similar formation schemes such as 4-2-3-1 and 4-3-3. 
For this reason, we have suggested a novel measurement to quantify the results for formation classification and visualization based on the similarity to pre-defined formation templates. 
The results demonstrated that our visual formation summary already provides valuable information and is capable to summarize individual scenes in football matches. In addition, we have shown the superiority of our classification approach compared to the current state of the art.

In the future, we plan to extend the current analytics system with other valuable tactical indicators such as the variance and movements of the players. Our current approach explicitly aimed for a solution that does not require any training data for classification. However, due to the increasing amount of position data, whether synthetic or real, deep learning approaches could become applicable to find more sophisticated solutions for the classification of team formations.



%

%
%
%
%
\bibliographystyle{ACM-Reference-Format}
\bibliography{references}

%

\end{document}